\documentclass[conference]{IEEEtran}
\IEEEoverridecommandlockouts
\usepackage{cite}
\usepackage{amsmath,amssymb,amsfonts}
\usepackage{algorithmic}
\usepackage{graphicx}
\usepackage{textcomp}
\usepackage{xcolor}
\def\BibTeX{{\rm B\kern-.05em{\sc i\kern-.025em b}\kern-.08em
    T\kern-.1667em\lower.7ex\hbox{E}\kern-.125emX}}
\begin{document}

\title{Carrot Cure: A CNN Based Application to Detect Carrot Disease\\
{\footnotesize \textsuperscript{}}

}
\author{\IEEEauthorblockN{Shree. Dolax Ray}
\IEEEauthorblockA{\textit{Department of CSE} \\
\textit{Daffodil International University}\\
Dhaka, Bangladesh \\
dolax15-9468@diu.edu.bd}
\and
\IEEEauthorblockN{Mst. Khadija Tul Kubra Natasha}
\IEEEauthorblockA{\textit{Department of CSE} \\
\textit{Daffodil International University}\\
Dhaka, Bangladesh \\
khadija15-10307@diu.edu.bd}
\and
\IEEEauthorblockN{Md. Azizul Hakim}
\IEEEauthorblockA{\textit{Department of CSE} \\
\textit{Daffodil International University}\\
Dhaka, Bangladesh \\
azizul.cse@diu.edu.bd}
\and
\IEEEauthorblockN{Fatema Nur}
\IEEEauthorblockA{\textit{Department of IT} \\
\textit{Monash University}\\
Subang Jaya, Malaysia \\
fnur0003@student.monash.edu}
\and
}

\maketitle


\begin{abstract}
Carrot is a famous nutritional vegetable and developed all over the world. Different diseases of Carrot has become a massive issue in the carrot production circle which leads to a tremendous effect on the economic growth in the agricultural sector. An automatic carrot disease detection system can help to identify malicious carrots and can provide a guide to cure carrot disease in an earlier stage, resulting in a less economical loss in the carrot production system. In this paper, we have developed a web application “Carrot Cure” based on Convolutional Neural Network (CNN) which can identify a defective carrot and provide a proper curative solution. Images of carrots affected by cavity spot and leaf bright as well as healthy images were collected. In this research, we've employed Convolutional Neural Network to include birth neural purposes and a Fully Convolutional Neural Network model (FCNN) for infection order. We've explored different avenues regarding different convolutional models with colorful layers and the proposed Convolutional model achieved the perfection of virtually 99.8\%, which is surely useful for the drovers to distinguish carrot illness and boost their advantage.
\end{abstract}

\section{Introduction}
To withstand the changing economic condition in Bangladesh, the agriculture industry requires a major upgrade. Agriculture provides work for 63\% population in Bangladesh's densely populated country \cite{b1}. A complex interplay of soil seed results in the agricultural production system. The taproot of the carrot is the most often consumed component of this root vegetable\cite{b2}. Fiber, beta carotene, vitamin K1, antioxidants, and potassium are all found in abundance in carrots. Beta-carotene is found in abundance in carrots. It is converted to vitamin A in the human body. Carrots include fiber that can help keep blood sugar levels in check. As crop disease refers to any detrimental departure or variation from the physiological systems' normal functioning. In current agricultural science, there are notable research possibilities and opportunities that are not being adequately developed. Because of rising population and political instability, the agriculture sectors began looking for new ways to enhance food production. In a successful farming system, disease diagnosis in crops is critical. Carrot infections are a serious issue in the agricultural industry. Carrots have a plethora of illnesses. One of the primary issues is cavity spot and leaf blight. A farmer detects disease signs in plants by observing them which necessitates constant monitoring. However, in big plantations, this method is more expensive and less precise. Farmers may utilize precision agricultural information technology to gather data and information to make informed decisions about how to maximize farm productivity. In this study, an automated method was built to assist farmers in detecting illnesses by collecting images with a camera and then entering the images into a processing system. To identify various carrot illnesses, a Deep learning approach is used. As a result, an accurate value and less expensive Machine Vision System are required to detect diseases from images and recommend the appropriate pesticide as a solution. We have divided our main goals into five major parts mentioned below:
\begin{itemize}
    \item To analyze the image distorted of carrots, our proposed method is very efficient to recognize the disease of the carrot.
    \item Exceeded the data set amount by doing some image process techniques, background noise removal, simulating data.
    \item Analyzed five different CNN architectures and evaluated the performance of the classification and identification of carrot diseases.
    \item The accuracy rate we have gained from this study surpasses the previous work.
    \item Our work will help the farmer to diagnose carrots and adopt timely treatment methods. As a result, the farmer brothers will also benefit from good harvest and economic aspects.
\end{itemize}
The rest of the part of this paper is organized as follows. Section II summarized the related works. Proposed methodology is described in section III. We have analysed the experimental evaluation in the Section IV. The proposed application architecture described in the section V. Finally we have concluded our work along with stating the future work in section VI.
\section{Related Works}
Several researchers have done researches on carrot disease detection as well as different fruits and vegetable diseases classification. In this part, we have mentioned some of them.\\
Rupali Saha et al.\cite{b3} the author published an exploration paper that an orange fruit complaint bracket. They used the deep learning algorithm which is Convolutional Neural Network (CNN). The algorithm used to identify the three conditions of orange. The dataset size was 68 where their system can detect 8 features. The accuracy is 93.21\% of their approach.\\
MO Al-Shawwa et al.\cite{b4} Apple fruit classification is performed on different kind of apples, by using Deep literacy for the bracket and discovery of the breed of an apple. They proposed a machine learning based approach to identify different types of apple. Out of their total dataset of images, they used 4,488 images for training purpose, for validation purpose 1,928, and for testing purpose they have used 2,138 images. To better train their deep learning system, they used 70\% images of the total dataset to train the model and 30\% images for validation. The model reaches 100\% accuracy on the test dataset.\\
TT Mim et al.\cite{b5} Sponge gourd complaint acknowledgment, for feting the splint and flower situation by exercising Convolutional Neural Network and image processing techniques. A pre-trained model used to detect the disease which is called Alexnet. The system takes some pictures as input and detect the conditions. The achieved delicacy is 81.52\%.\\
Y. N. P. H. H. Zaw Min Khaing et al.\cite{b6} the author has created a control system for expulsion. Where the use of a CNN-based recognition is considered. The system is applied to fruit discovery and recognition through parameter optimization. Their results were 94\% of the system.\\
Abeer A. Elsharif et al.\cite{b7} an exploration paper published the dataset used approximately 2400 images of potatoes which was a public dataset. They trained a CNN model which can identify four types of potatoes. The trained model can give 99.5\% accuracy, demonstrating the feasibility of this method.\\
G. C. Khadabadi et al.\cite{b8} the author used image processing techniques ranging from image acquisition, pre-processing. Use the Discrete Wavelet Transform (DWT) system to diagnose and classify carrot vegetables. The accuracy of their proposed system is 88.0\% accuracy.\\
Methun NR et al.\cite{b9} in their research, they used CNN Architecture to detect total of five diseases of the carrot, including healthy carrots. Where they have used and tested four more different pre-trained models which are viz, VGG16, VGG19, MobileNet, and Inception V3. They trained and evaluated the system, using a powerful dataset, consisting of original and synthetic data. They got 97.4\% accuracy using Inception v3.\\
H. Zhu et al.\cite{b10} they have established a system based on a deep learning structure. Carrot images are tested by preparing the identification model on the Alexnet network, pre-trained by a large computer vision database (image-net). Which trains neural networks with less data than traditional CNN. Applying this method to the carrot image data set, the accuracy was 98.70\%.\\
P. Ganesh et al.\cite{b11} Get a pixel-based mask to identify each fruit and for each marked fruit in an image. Advanced Segmentation Framework system, R-CNN-Based, and Pixel-Based fruit identification. Oranges have been identified using a deep learning method. The algorithm's performance is compared between the RGB and RGB + HSV images. Their first results show that, after the insertion of data, the accuracy increases from 0.8947 to 0.9753. When the RGB data alone is used. They got the F1 score accuracy is 0.89\%.\\
M. T. Habib et al.\cite{b12} the author proposed K-means clustering algorithms to divide disease-infested areas from taken images. Then the properties needed to classify diseases are extracted using a support vector machine. The accuracy is 90\% of their approach.\\
L. J. Rozario et al.\cite{b13} they have developed a method by which they can diagnose or identify fruits and vegetables. Four categories of fruits and vegetables can be identified through their system. For the color-based segmentation of images they have used Pre-modified K-mean clustering and Otsu methods. They used a small size of the dataset, 63 pictures in this work. This work has not been classification.\\
S. A. Gaikwad et al.\cite{b14} they build a system which can detect and classify fruit disease using image processing. They used a K-mean clustering algorithm for image segmentation. Their process include features extraction, split image and finally used an SVM classification for classification.\\
M. Islam et al. in\cite{b15} author proposed a compact approach to merge image processing and machine learning techniques for detecting the potato tree malady. The area with the symptoms in the pictures was drawn in L * a * b * colored spots. Here, total 10 features can detect. The dataset was more than 300 images, they applied the multiclass SVM to classify diseases of the potato image. The accuracy gets 95\%.\\
S. K. Behera et al.\cite{b16} proposed a system that has the ability to identify a distortion of orange fruit. Also, the type of disease that can be seen on the surface of the orange fruit. It can identify. Defective signs indicate the severity of the disease and recommend the best way to deal with the disease. Multi-class SVM with K-means clustering is used for the classification of diseases. They got the accuracy of their system is 90\%.\\
M. Dhakate et al.\cite{b17} they use neural networks to diagnose pomegranate trees. Image pre-processing and k-means are used to create clustering segmentation systems. Textured properties are extracted using the GLCM method and given to artificial neural networks. In this method, the accuracy is 90\%.\\
S. M. Farhan Al Haque et al.\cite{b8} the publisher have proposed a system that can detect disease of the guava and cure of disease which is a CNN model. They have collected their dataset from the different districts of Bangladesh. They experiment with three CNN models. After that experiment, the experimental result saw that their third model one was 95.61\% which is more accurate than the other two models.\\
After comparatively analyzing the previous works, it is found that different researchers have applied different methods to classify carrot images as well as other fruits and vegetables. We aim to predict carrot disease as well as differentiate between fresh and defective carrot images that can help our farmers and agriculturists to increase the production of carrots.
\section{Methodology}
\subsection{Data Collection and Utilization}
\begin{figure}[htbp]
  \centering
  \begin{tabular}{c c c}
  \includegraphics[width=0.9in]{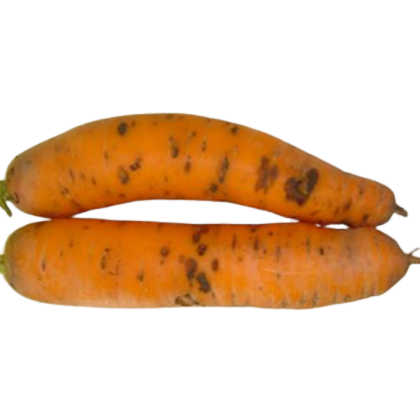} &
  \includegraphics[width=0.7in]{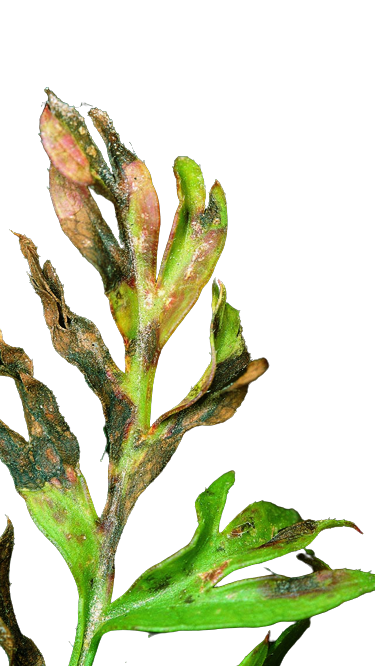}&
  \includegraphics[width=1.0in]{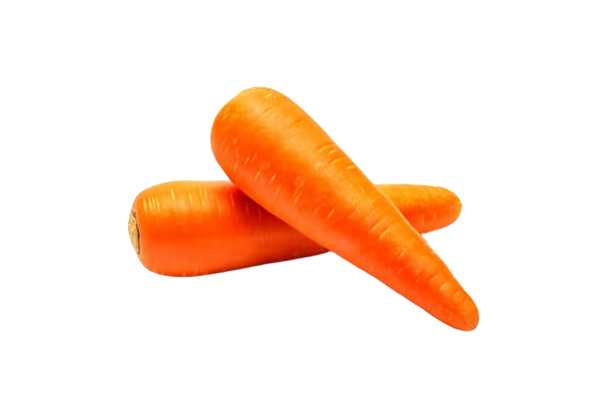}\\
  a.~Cavity Spot & b.~Leaf Blight & c.~Healthy
  \end{tabular}
  \caption{Images of Carrot Diseases \& Healthy Carrot}
  \label{fig1}
\end{figure}

In this exploration, we've used a dataset for image processing. We've collected most of the images from different districts of Bangladesh and some images were collected from internet sources as well. Firstly, we've collected raw images per complaint and have increased the dataset by using image augmentation techniques such as Rotate, Shear, Width-shift, Height-shift, and Horizontal-flip. As shown in ``Fig.~\ref{fig1}''.
\subsection{Data Pre-Processing}
As we've collected the images from different sources there was noise in the dataset. This was one of the vital challenges to remove those noises. Since the classification results are mostly dependent on the condition of the images, it is necessary to minimize the noise from the images to make a standard dataset for the work. In our work, we have used Fuzzy Filter\cite{b30} to reduce the Gaussian noise. The main reason to use this filter is that we can minimize the noise from the images without hampering the other features. We've put our dataset into four classes are Cavity Spot, Leaf Blight, Healthy, and Fresh Carrot, as shown in ``Table.~\ref{table1}''. The dataset is divided into two portions which are the training set and the alternate bone is validation set.\\
\begin{table}[htbp]
 \caption{Dataset table}
\label{table1}
\centering
\begin{tabular}{|c|c|c|}
\hline
No.             & Class Name    & Train Data    \\ \hline
1               & Cavity Spot   & 456           \\ \hline
2               & Healthy       & 205           \\ \hline
3               & Leaf Blight   & 264           \\ \hline
4               & Fresh Carrot  & 148           \\ \hline
Total: 4 Class  &               &1063           \\ \hline
\end{tabular}
\end{table}

\subsection{Proposed Workflow}
\begin{figure}[htbp]
\centerline{\includegraphics[width=0.45\textwidth]{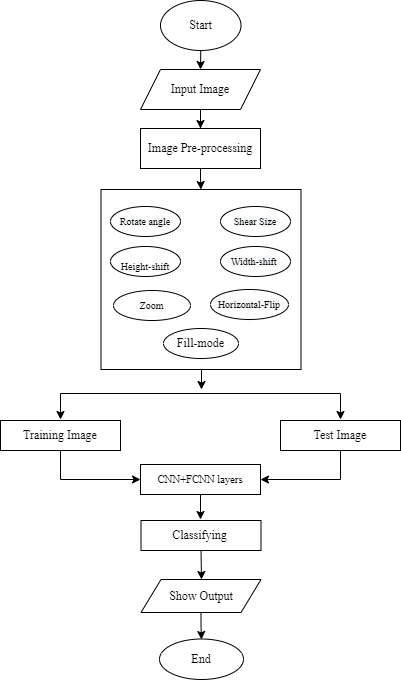}}
\caption{Proposed Workflow Model}
\label{fig2}
\end{figure}

The system starts with providing the images of carrots into the processing unit, segmentation unit, point birth unit, training unit, and so on stated as below in ``Fig.~\ref{fig2}''.\\
Step 1-Data collection: we’ve collected the major portion of images from different districts of Bangladesh and some minor portions of images were collected from different internet sources. \\ 
Step 2-Data pre-processing unit: We've removed the noise, background of the collected images and have prepared synthetic data. Finally, we've increased our dataset by using different Data Augmentation\cite{b19} techniques such as:
\begin{itemize}
    \item Rotate
    \item Width-shift 
    \item Height-shift
    \item Shear
    \item Zoom
    \item Horizontal-flip 
    \item Fill-mode
\end{itemize}
Step 3-Resizing the Data: The collected raw images were in different sizes. So, we've to bring them in a similar size and in that case, we've resized them all.\\
Step 4-Splitting the Data: Training data creators to train and validate our data for better delicacy we've named some models. To get better delicacy with our machine configuration, we've enforced five different models and incipiently, one model has been named for the final training and testing process.\\
Step 5- Evaluating the Pre: In this part, we have evaluated each model using different performance evaluation metrics.\\
Posterior to training, validation and testing that commerce has provided us many rigors graph with training and confirmation delicacy \& training and confirmation loss. Also, we've calculated the confusion matrix and a table for understanding the perfection, recall, and f1 score.
\subsection{Proposed Methodology}
\begin{figure*}[htbp]
\centerline{\includegraphics[width=0.9\textwidth]{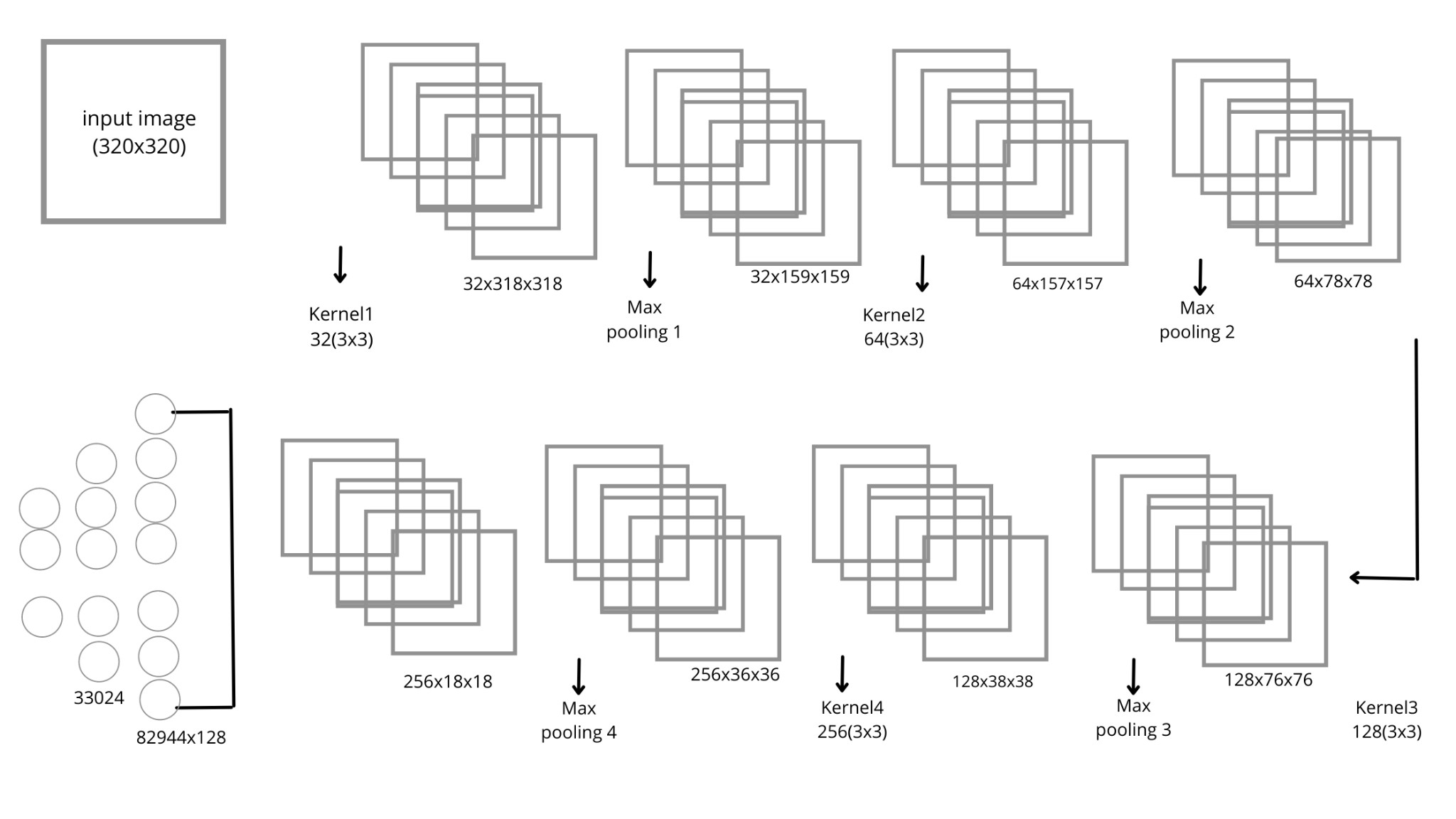}}
\caption{Proposed CNN layer.}
\label{fig3}
\end{figure*}

We have proposed a Convolutional Neural Network, which has 4 convolutional layers, 2*2 max-pooling layers, and 2*2 dense layers . As shown in ``Fig.~\ref{fig3}''.\\
The first convolutional layer contains 32-3x3 Filters, and the activation function is "Relu" with Max pooling layer of (2x2).\\
The second layer consists of 64-3x3 Filters, and the activation function is "Relu" where the Max pooling layer was (2x2).\\
The  configuration of the third layer was 128-3x3 Filters, and the activation function is "Relu" and Max pooling (2x2).\\
The fourth layer consists of 256-3x3 Filters, and the activation function is "Relu" with Max pooling (2x2) and Dropout (0.5) [24].\\
First Dense units: 128, and the activation function is "Relu". 
Dropout (0.25).\\ 
Second Dense units: 256, and the activation function is "softmax".
\subsection{Software and Hardware Requirements}
After the appropriate examination of all important measurable or theoretical ideas and methods, a list of necessities has been created that should be needed for such a work of Classification. The minimum Hardware and Software requirements are:
\begin{itemize}
    \item Operating system (Windows 7 or above)
    \item Minimum 100 GB Hard-disk
    \item 4GB RAM 
    \item Python Environment
    \item Google Colab
\end{itemize}
\section{Results and Analysis}
In this research work, we have trained the dataset among five models with different datasets. One dataset has 506 images and another dataset has 1063 images. Different metrics can be used to measure the quality of machine learning models. Evaluation metrics\cite{b28} are defined as follows:
\begin{align*}
  Accuracy      &= {\frac{T_P + T_N}{T_P + F_N + F_P + T_N}} \times 100\% \\
  Recall        &=  {\frac{T_P}{T_P + F_N}}\times 100\%\\
  Specificity   &=  {\frac{T_N}{T_N + F_P}}\times 100\%\\
  Precision     &=  {\frac{T_P}{T_P + F_P}}\times 100\%\\
  F1-score      &=  \frac{2 \times Precision \times Recall}{Precision + Recall}
\end{align*}
Where, $T_P$, $T_N$, $F_P$, and $F_N$ represent True Positive, True Negative, False Positive, and False Negative\cite{b20}.\\
In the first model, we have used Convolutional Neural Network (CNN) where five Max-Pooling layers, and three dense layers. The accuracy we have gotten is 98.40\%. ``Table.~\ref{table2}'' shows the performance evaluation of the first CNN Model.
\begin{table}[htbp]
 \caption{Performance Evaluation of the CNN Model}
\label{table2}
\centering
\begin{tabular}{|c|c|c|c|}
\hline
Disease/Class Name    & Precision    & Recall   &F1 Score   \\ \hline
Cavity Spot           & 99.5\%       & 98.8\%   & 99.15\%   \\ \hline
Leaf Blight           & 99.0\%       & 98.5\%   & 98.74\%   \\ \hline
Fresh Carrot          & 98.0\%       & 97.0\%   & 97.49\%   \\ \hline
Fresh Leaf            & 100\%        & 98.0\%   & 98.99\%   \\ \hline
\end{tabular}
\end{table}

\begin{figure}[htbp]
\centerline{\includegraphics[width=0.4\textwidth]{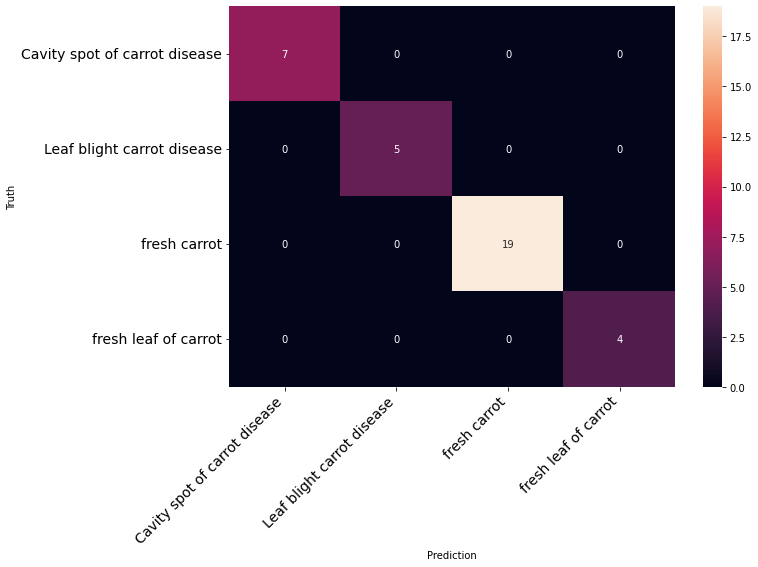}}
\caption{Confusion Matrix for the Proposed CNN Model}
\label{fig4}
\end{figure}

In the second model, we have used three max-pooling layers and one dense layer. The accuracy we have gotten is 99.61\%.\\
Then in the third model, we have used four max-pooling layers and one dense layer. The third accuracy is 97.40\%.\\
We have used two max-pooling layers and one dense layer in the fourth model and this model achieved an accuracy of 99.01\%.\\
The fifth and final model has four Max- Pooling layers and two dense layers which have given the best accuracy. In the Fifth model, we try to optimize our CNN model and finally get the best accuracy of 99.81\%. ``Fig.~\ref{fig4}'' depicts the confusion matrix for the fifth model.

\begin{figure}[htbp]
\centerline{\includegraphics[width=0.45\textwidth]{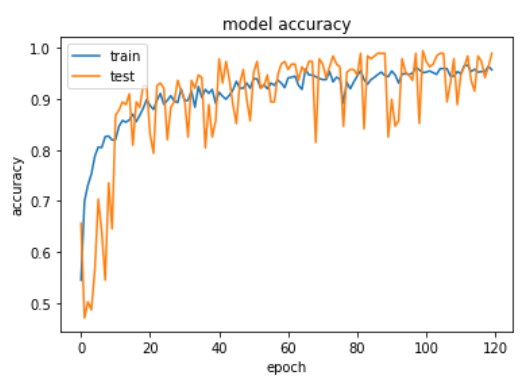}}
\caption{Model accuracy (4 max-pooling and 2 dense).}
\label{fig5}
\end{figure}

\begin{figure}[htbp]
\centerline{\includegraphics[width=0.45\textwidth]{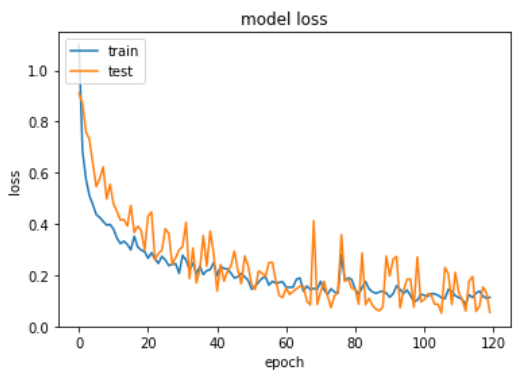}}
\caption{Model loss (4 max-pooling and 2 dense).}
\label{fig6}
\end{figure}

The validation accuracy and validation loss for the fifth model shows in the ``Fig.~\ref{fig5}'' and ``Fig.~\ref{fig6}''. The architectures and validation accuracy for all the five models are shown in ``Table.~\ref{table3}''.

\begin{table}[htbp]
\caption{Models and validation accuracy}
\label{table3}
\centering
\begin{tabular}{|c|c|c|c|}
\hline
Model Number &Max-pooling Layer &Dense Layer &\begin{tabular}[c]{@{}l@{}}Validation\\Accuracy \end{tabular}\\ \hline
First model  &5       &3   &98.40\%   \\ \hline
Second model &3       &1   &99.61\%   \\ \hline
Third model  &4       &1   &97.40\%   \\ \hline
Fourth model &2       &1   &99.01\%   \\ \hline
Fifth model  &4       &2   &99.81\%   \\ \hline
\end{tabular}
\end{table}

In the final model we try to optimize our Convolutional Neural Network (CNN) model and increase data, then we get the best accuracy of our model is 99.81\%.

\begin{table*}[htbp]
\caption{Comparison with other works}
\label{table4}
\centering
\begin{tabular}{|l|l|l|l|l|l|l|l|}
\hline
~~ Reference & \begin{tabular}[c]{@{}l@{}}Problem Domain\end{tabular} & \begin{tabular}[c]{@{}l@{}}Object\end{tabular} &\begin{tabular}[c]{@{}l@{}}Data Size\end{tabular} &\begin{tabular}[c]{@{}l@{}} Algorithm/Method\end{tabular} &\begin{tabular}[c]{@{}l@{}} Classifier\end{tabular} &\begin{tabular}[c]{@{}l@{}} No of object\\detectable\end{tabular} &\begin{tabular}[c]{@{}l@{}} Accuracy(\%)\end{tabular} \\ \hline

This Work & \begin{tabular}[c]{@{}l@{}}Disease Detection and\\ Classification\end{tabular} & Carrot & 1467 & Deep Learning & CNN & N/A  & 99.81 \\ \hline

Rupali Saha et al.\cite{b3}& \begin{tabular}[c]{@{}l@{}}Disease Detection and\\ Classification\end{tabular} &Orange &68 &\begin{tabular}[c]{@{}l@{}}Color threshold \\segmentation\end{tabular} &CNN &8 &93.21 \\ \hline

G. C. Khadabadi et al.\cite{b8}& \begin{tabular}[c]{@{}l@{}}Disease Detection and\\ Classification\end{tabular} &Carrot &50 &Not mentioned &PNN &4 &88 \\ \hline

NR Methun et al.\cite{b9}& \begin{tabular}[c]{@{}l@{}}Disease Detection and\\ Classification\end{tabular} &Carrot &2131 &Not mentioned &CNN &Not mentioned &97.4 \\ \hline

M. T. Habib et al.\cite{b12}& \begin{tabular}[c]{@{}l@{}}Disease Detection and\\ Classification\end{tabular} &Papaya &129 &k-means clustering &SVM &10 &90.15 \\ \hline

S. Sasirekha et al.\cite{b21}& Disease Recognition &Carrot &Not mentioned &k-means clustering &\begin{tabular}[c]{@{}l@{}}Multiclass\\SVM\end{tabular} &13 &\begin{tabular}[c]{@{}l@{}}Not\\mentioned\end{tabular} \\ \hline

B. J. Samajpati et al.\cite{b22}& \begin{tabular}[c]{@{}l@{}}Disease Detection and\\ Classification\end{tabular} &Apple &80 &k-means clustering &\begin{tabular}[c]{@{}l@{}}Random\\Forest\end{tabular} &13 &60-100 \\ \hline
\end{tabular}
\end{table*}

\section{Application Architecture}
\subsection{Application Feature}
\begin{figure}[htbp]
\centerline{\includegraphics[width=0.45\textwidth]{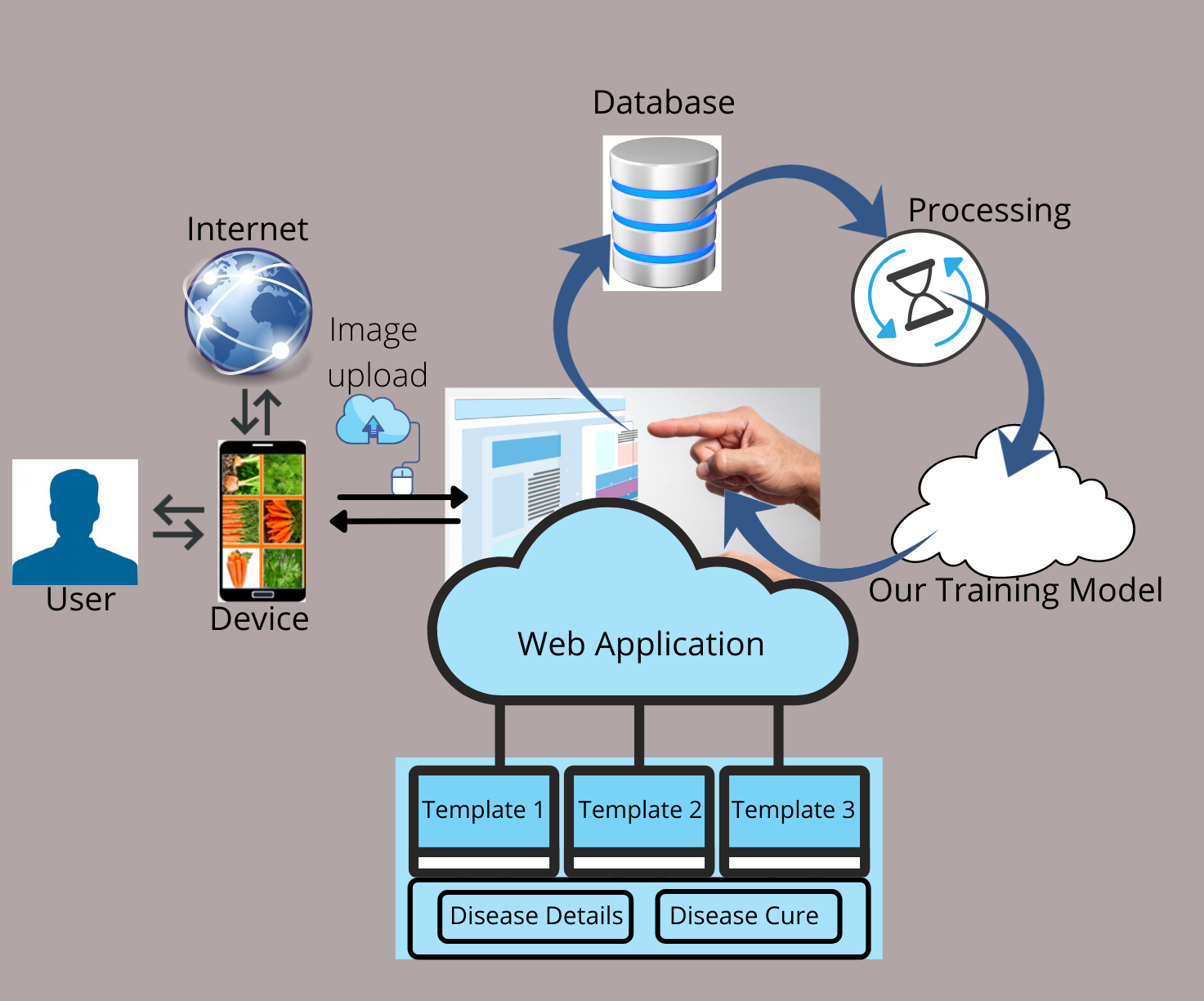}}
\caption{Carrot Cure Application Architecture.}
\label{fig7}
\end{figure}

We are living in a modern world where the use of artificial intelligence systems is increasing day by day. Our AI system will help to predict carrot disease which will easier to find or detect carrot disease\cite{b24}. Deep learning and AI are a blessing for our world. If we use it properly, we can do a lot better, today is a small example of where we will see what happens in carrot cultivation. Learn about the disease, our application feature given in ``Fig.~\ref{fig7}''. First of all our system will require an image\cite{b25}. Then the image will process with our model and it will show the output\cite{b26}. A landing page will show to the user. The landing page will show based on our model result. If the user image will match our model. Then the user directly will go to the template and the template will show the user what will next step for him.The software has the following features:
\begin{itemize}
    \item It can detect carrot disease.
    \item Diseases in English and Bengali will be given by the application. 
    \item Diseases will be cured in English and Bengali on the application.
    \item The application will be able to give the name of the disease drug.
\end{itemize}
\subsection{Technical Requirement}
To use this application, we need to use any smartphone (android or IOS) or a computer and must have a browser which can be any browser Microsoft Edge, Google Chrome, Opera Mini, UC Browser, Mozilla Firefox, etc.\cite{b28}. The application will work better or smooth on android version 4.4 KitKat or above, IOS version 8 or above, and the desktop version windows 7 minimum required with the internet. The insert images will be captured using a camera. The resulting enhancements are displayed on the display panel.

\subsection{Software development kit}
We used Google Colab, Pycharm, Python, HTML, CSS, bootstrap to create our application. We create this web application by a laptop windows 10 pro, 8 GB ram, 256 GB, core i5 processor SSD.
\section{Conclusion And Future Work}
This research proposed a system which will identify unhealthy carrots which will assist the farmers and agriculturists to cultivate fresh carrots. The main motivation of our work is to identify the factors which will help those growers so that they can fluently get their awaiting crop details through our model and can take the necessary steps. Pre-processing techniques were employed to remove noise. Well processed images were trained and tested by using CNN models. Finally, we've got the stylish training and confirmation delicacy graph of our design. Eventually, we've planted our anticipated result which is that the delicacy is stylish and it's 0.99812\%. By this model, we have created a web application. The application can do detect carrot diseases. It can show more details about the disease and give the solution or cure. It also has features like medicine recommendations. In the future, the number of conditions the system identifies could be better. And complaint inflexibility can also be linked.

\vspace{12pt}

\end{document}